\documentclass[letterpaper]{article} 
\usepackage{aaai24}  
\usepackage{times}  
\usepackage{helvet}  
\usepackage{courier}  
\usepackage[hyphens]{url}  
\usepackage{graphicx} 
\usepackage{natbib}  
\usepackage{caption} 
\usepackage{algorithm}
\usepackage{algorithmic}

\usepackage{newfloat}
\usepackage{listings}
\DeclareCaptionStyle{ruled}{labelfont=normalfont,labelsep=colon,strut=off} 
\floatstyle{ruled}
\newfloat{listing}{tb}{lst}{}
\floatname{listing}{Listing}
\usepackage{verbatim}
\usepackage{amssymb}
\usepackage{caption}
\usepackage{subcaption}
\usepackage[labelformat=simple]{subcaption}

\usepackage{bm} 
\usepackage{amsmath}
\usepackage{kotex}
\usepackage{multirow}

\usepackage[export]{adjustbox}

\usepackage[utf8]{inputenc}
\usepackage{dcolumn} 
\usepackage{multirow}
\usepackage{adjustbox}
\usepackage{booktabs,caption}
\usepackage[flushleft]{threeparttable} 
\newcommand\mc[1]{\multicolumn{1}{c}{#1}} 

\usepackage{pifont}

\title{Domain Generalization with Vital Phase Augmentation}
\author{
    Ingyun Lee, Wooju Lee, Hyun Myung\thanks{The corresponding author.}
}
\affiliations{
    Urban Robotics Lab, School of Electrical Engineering, \\Korea Advanced Institute of Science and Technology, Republic of Korea\\

    \{iglee, dnwn24, hmyung\}@kaist.ac.kr
}

\begin{document}

\maketitle

\begin{abstract}
Deep neural networks have shown remarkable performance in image classification. However, their performance significantly deteriorates with corrupted input data. Domain generalization methods have been proposed to train robust models against out-of-distribution data. Data augmentation in the frequency domain is one of such approaches that enable a model to learn phase features to establish domain-invariant representations. This approach changes the amplitudes of the input data while preserving the phases. However, using fixed phases leads to susceptibility to phase fluctuations because amplitudes and phase fluctuations commonly occur in out-of-distribution. In this study, to address this problem, we introduce an approach using finite variation of the phases of input data rather than maintaining fixed phases. Based on the assumption that the degree of domain-invariant features varies for each phase, we propose a method to distinguish phases based on this degree. In addition, we propose a method called vital phase augmentation (VIPAug) that applies the variation to the phases differently according to the degree of domain-invariant features of given phases. The model depends more on the vital phases that contain more domain-invariant features for attaining robustness to amplitude and phase fluctuations. We present experimental evaluations of our proposed approach, which exhibited improved performance for both clean and corrupted data. VIPAug achieved SOTA performance on the benchmark CIFAR-10 and CIFAR-100 datasets, as well as near-SOTA performance on the ImageNet-100 and ImageNet datasets. Our code is available at https://github.com/excitedkid/vipaug.
\end{abstract}

\section{Introduction}
\begin{figure}[t!]
    \begin{center}
       \includegraphics[width=0.4\textwidth]{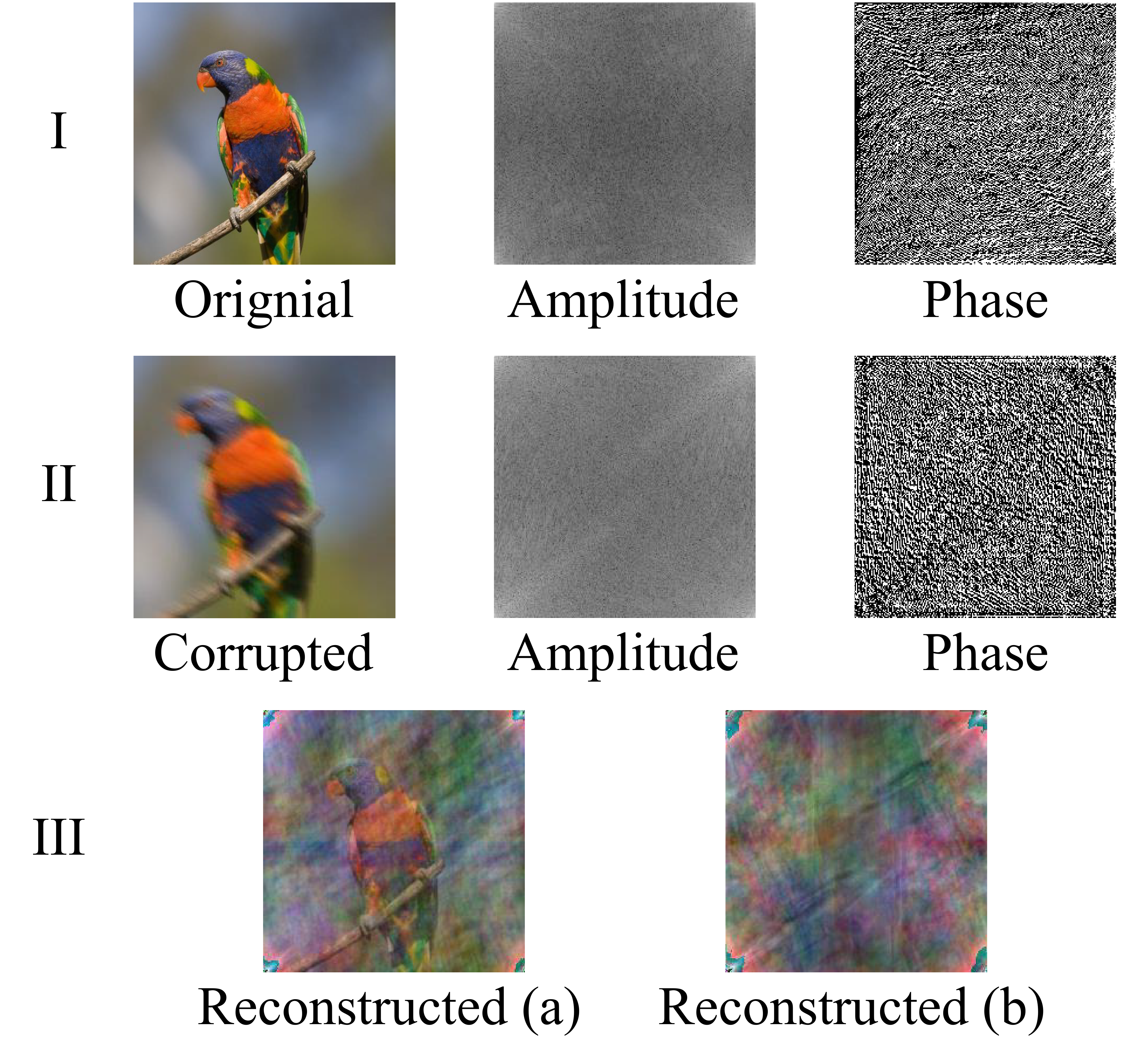}
    \end{center}
    \caption{Amplitude and phase of original and blur images, with reconstructed images. Amplitude and phase fluctuations can be observed in \uppercase\expandafter{\romannumeral2} compared with \uppercase\expandafter{\romannumeral1}. \uppercase\expandafter{\romannumeral3}-(a) retains vital phases of the original image and sets other phases to 0. In contrast, \uppercase\expandafter{\romannumeral3}-(b) retains the non-vital phases at the same ratio and sets other phases to 0. In \uppercase\expandafter{\romannumeral3}, the amplitude of the original image is kept unchanged. The amplitude and non-vital phases are shown to contain less domain-invariant features.
    } 
    \label{figure1}
\end{figure}
Deep learning is being actively explored for various applications in computer vision such as image classification and object detection~\cite{he2016deep28, tan2019efficientnet, dosovitskiy2021an}. The rapid development of deep learning methods has led to performance that can surpass that of human effort on some tasks. For example, deep neural networks (DNNs) can achieve high accuracy on image classification tasks with in-distribution data. However, the real-world performance of DNNs can be poor compared with that of manual classification by humans~\cite{hendrycks2019benchmarking27}. Because the distributions of train and test datasets may differ in the real world, deep learning models cannot be trained to compensate for all of the potential types of data corruption. To address this challenge, domain generalization methods have been developed to train models to be more robust to out-of-distribution (OOD) data~\cite{lee2022adversarial}. These techniques aim to minimize any deterioration in performance on clean data while improving the performance of deep learning models on corrupted data.

\begin{figure}[!tbp]
\centering
\includegraphics[width=0.85\columnwidth, ]{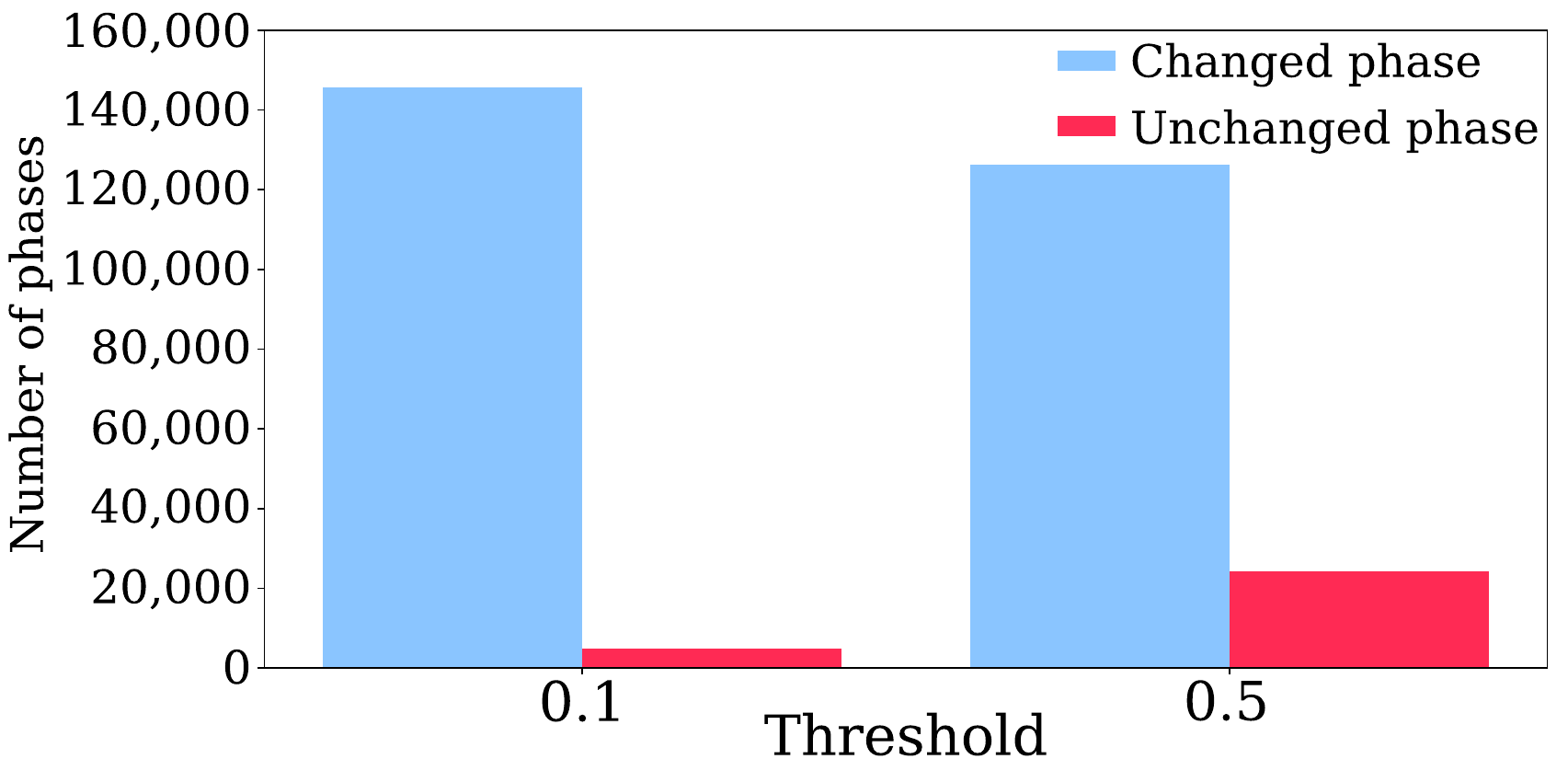}
\caption{The number of fluctuated phases from clean to corrupted domains. We extracted an arbitrary image from ImageNet-C and calculated the average value of all corruption types at corruption severity level 3. Phase fluctuations above the threshold value were counted. The range of phase is $[-\pi, \pi]$. Most phases were changed even when the threshold was small.}
\label{figure2}
\end{figure}

Data augmentation methods have also been proposed to improve domain generalization. Some of these approaches~\cite{chen2021amplitude15,xu2021fourier16} based on the frequency domain show that phases contain domain-invariant features. To make the models depend on the phases, only the amplitudes of the input data are varied with several different techniques and fix the phases. The fixed phases are then combined with the augmented amplitudes to reconstruct the image. However, with corrupted data, amplitudes and phases can fluctuate significantly as shown in Figure~\ref{figure1} and Figure~\ref{figure2}. Therefore, existing methods with fixed phases are not robust to phase fluctuations.

To address the limitations of existing methods, we propose to introduce finite phase variations to ensure robustness to phase fluctuations. We propose two hypotheses. First, the degree of domain-invariant feature inclusion, which we define as the robustness weight, varies for each phase. We define a phase with relatively high robustness weights as a vital phase and a phase with low robustness weights as a non-vital phase. Accordingly, we propose a method to detect vital and non-vital phases based on the magnitude of the amplitudes. Second, applying different strengths of variations according to robustness weights allows a model to depend more on vital phases, which enhances its robustness against corruption. By retaining the advantages of existing methods and addressing the vulnerability to phase fluctuations, we propose a novel approach called vital phase augmentation (VIPAug).

VIPAug applies variations to the phases of input data based on robustness weight and replaces all amplitudes, enabling the model to depend on the vital phases. VIPAug incorporates phase variations by employing a Gaussian distribution and partially replacing the phases with those of fractal images. We also present the experimental results of our approach, which show the improved accuracy on both clean and corrupted data compared with baseline methods. The contributions of this study are summarized as follows:
\begin{itemize} 
\item We experimentally demonstrate that the robustness weights of phases differ for the first time.
\item We propose a method to identify vital and non-vital phases based on their weights. 
\item We propose VIPAug as a novel augmentation approach that combines the new phase variations with existing methods based on amplitudes variations. This approach enables the model to perform more robustly against phase fluctuations while depending on the phases. 
\item Our experimental results show that the proposed method achieved state-of-the-art performance on the CIFAR-10 and CIFAR-100 datasets~\cite{krizhevsky2009cifar25} and nearly state-of-the-art performance on the ImageNet-100 and ImageNet datasets~\cite{deng2009imagenet26}.
\end{itemize}

\section{Related Work}
\subsection{Domain Generalization}

Deep learning models should be robust against unseen domains that may be used in real-world applications. Domain generalization methods aim to generalize models to OOD data by using only training data from a given source domain. Domain generalization can be implemented in a variety of ways, including contrastive learning, ensemble learning, and meta-learning. Contrastive learning methods reduce the multi-domain gap to improve generalization ability. Motiian~et~al.~\shortcite{motiian2017unified2} exploited the Siamese architecture with a contrastive loss. Yoon~et~al.~\shortcite{yoon2019generalizable3} extended a contrastive semantic alignment loss to mitigate the bias of data and establish domain-invariant representations. 

Ensemble learning methods combine several models to improve generalization. Ding~et~al.~\shortcite{ding2017deep5} used multiple domain-specific deep neural networks to capture a shared representation within multiple sources. Similarly, Liu~et~al.~\shortcite{liu2020ms6} proposed a multi-site network with domain-specific batch normalization layers. 

The meta-learning approach diversifies different models to improve their stability and generalization performance. Zhao~et~al.~\shortcite{zhao2021learning10} proposed a memory-based identification loss designed to harmonize with meta-learning. All these methods have a limitation in that they do not directly increase the diversity of the training data. For this reason, we concentrate on data augmentation among various methods for domain generalization.

\subsection{Data Augmentation}

Data augmentation has been studied to improve the generalization performance of deep learning models. Mixup~\cite{mixup19} is designed to mix two images with linear combinations to improve generalization ability. Cutout~\cite{cutout20} and Random Erasing~\cite{zhong2020random11} randomly erases a part of an image to improve accuracy and generalize to the occluded objects. AutoAugment~\cite{autoaugment13} optimizes a group of augmentations with reinforcement learning. However, these methods only generalize a model to limited scenarios and are not robust to various distributional shifts such as common corruptions.

\begin{figure*}[t!]
    \begin{center}
       \includegraphics[width=0.9\linewidth]{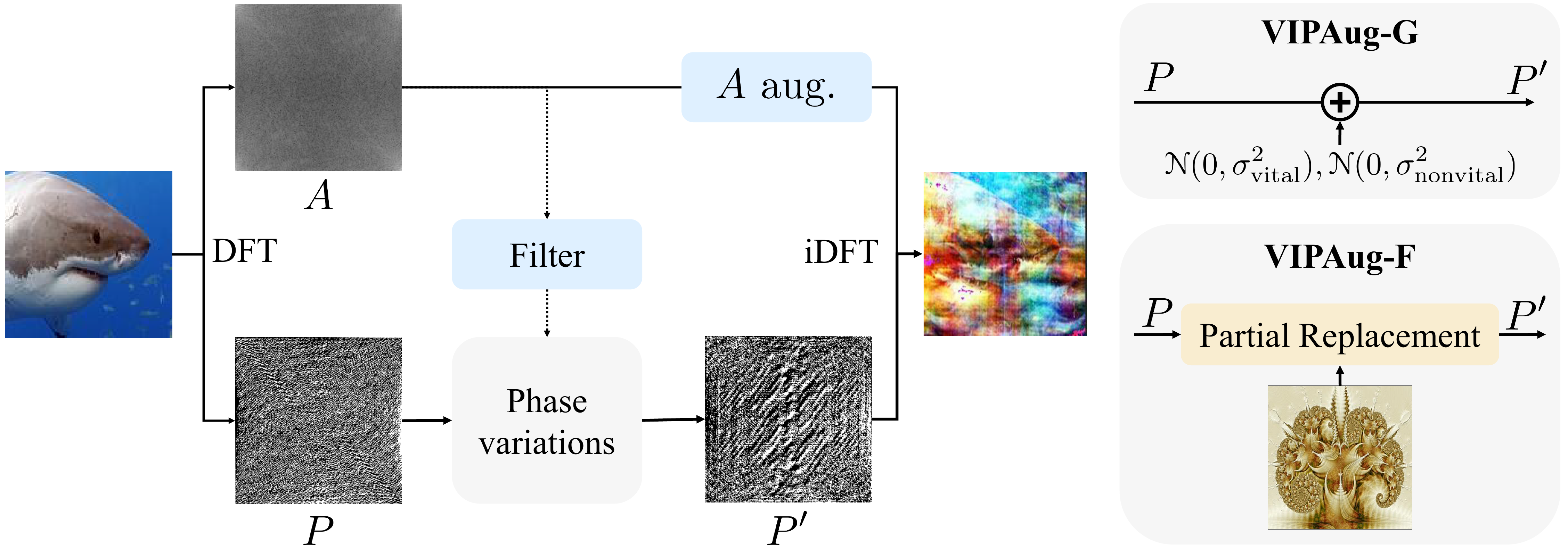}
    \end{center}
    \caption{Overall structure of VIPAug. VIPAug contains VIPAug-G and VIPAug-F. VIPAug-G introduces phase variations by using Gaussian distributions with different variances, $\sigma_{\text {vital }}^2$ and $\sigma_{\text {nonvital }}^2$. VIPAug-F employs the phases of the fractal images. We introduce finite phase variations by finding vital phases with a filter and applying variations with different strengths depending on the robustness weight. $A$, $P$, and $P^{\prime}$ denote the amplitude, phase, and varied phase spectrum.}
    \label{figure3}
\end{figure*}

Common corruptions refer to the possible distortions and distributional shifts in the real world such as shot noise, motion blur, and snow. Recently, several data augmentation methods have been proposed to improve performance on common corruption scenarios. These methods generate multi-source domains from a single-source domain using various transformations and mixing strategies. AugMix~\cite{augmix17} proposed parallel data pipelines to generate diverse domains while maintaining semantic content. PixMix~\cite{hendrycks2022pixmix14} mixes original images with external fractal images to introduce greater structural complexity. Zhou~et~al.~\shortcite{mixstyle12} randomly mixed instance-level feature statistics of training samples across source domains. However, these methods do not take into account that the image phases contain domain-invariant features.

Data augmentation using the frequency domain has also become a topic of active research that leverages domain-invariant features in images. APR-SP~\cite{chen2021amplitude15} fixes the phases and replaces the amplitudes with those from other images. FACT~\cite{xu2021fourier16} fixes the phases and mixes the amplitudes with those from other images. Both methods introduce amplitude variations to enable a model to learn domain-invariant features from the phases. However, fixing the phases makes DNNs vulnerable to phase fluctuations. HybridAugment++~\cite{hybridaug2023} makes the model rely on the low-frequency components of data, but this approach does not consider variations in the robustness weight of each phase. PRIME~\cite{modas2022prime18} is an integrated method that considered augmentation in the spectral, spatial, and color domains. Although this approach greatly increased diversity with augmentation in the three domains, the authors did not consider that the phase contains domain-invariant features in an image.

\section{Method}
We propose VIPAug as a data augmentation method that integrates changes in the amplitudes and finite variations in the phase spectrum. The phase variations apply different intensities of variations to vital and non-vital phases according to their robustness weights. First, we propose a method to distinguish the vital and non-vital phases using the magnitude of the amplitudes. VIPAug contains two types of phase augmentation; one utilizes Gaussian distributions, and the other employs fractal images. VIPAug encourages the model to depend on the phases over the amplitudes, specifically on the vital phases. Due to this dependence on the vital phases, the model achieves robustness against fluctuations in terms of both amplitudes and phases. The entire VIPAug process is shown in Figure~\ref{figure3}.  

\subsection{Detection of Vital Phase}
The conventional approach uses 2D discrete Fourier transform (DFT) for each channel of an RGB image to obtain amplitudes and phases. Unlike 2D DFT, 3D DFT can be used to acquire amplitudes and phases that include features between each channel. Leveraging these amplitude and phase spectrums can improve the accuracy on clean and corrupted data. With image's height $H$, width $W$, channel $C$, coordinates of image's spatial domain $(x,y,z)$ and frequency domain $(u,v,w)$, the 3D DFT equation is represented as follows: 
\begin{equation}    
F(u, v, w)=\sum_{x=0}^{H-1} \sum_{y=0}^{W-1} \sum_{z=0}^{C-1} f e^{-j 2 \pi\left(\frac{x}{H} u+\frac{y}{W} v+\frac{z}{C} w\right)},
\end{equation}
where the input image is represented by $f=f(x, y, z)$.
We can derive the image's amplitudes $A(u,v,w)$ and phases $P(u,v,w)$:
\begin{equation}  
\begin{aligned}
F(u, v, w) & =|F(u, v, w)| e^{j \cdot \arctan \frac{I(u, v, w)}{R(u, v, w)}} \\
& =A(u, v, w) e^{j P(u, v, w)},
\end{aligned}
\end{equation}\\
where $I(u,v,w)$ and $R(u,v,w)$ represent the imaginary and real parts of the DFT result. The relation between the image $f$ and its corresponding amplitudes $A(u, v, w)$ and phases $P(u,v,w)$ is described using inverse discrete Fourier transform (iDFT):
\begin{equation}
\begin{aligned}
& f=\frac{1}{H W C} \sum_{u=0}^{H-1} \sum_{v=0}^{W-1} \sum_{w=0}^{C-1} A e^{j\left\{2 \pi\left(\frac{u}{H} x+\frac{v}{W} y+\frac{w}{C} z\right)+P\right\}} ,
\end{aligned}
\end{equation}
where $A=A(u,v,w)$ and $P=P(u,v,w)$.
The image $f$ can be represented as a linear combination of complex exponential terms. The amplitude of the exponential term is proportional to the number of object features. Object features are semantically preserved across domains. Therefore, the phase containing more domain-invariant features has a larger amplitude. We hypothesize that the model should depend more on phases at larger amplitudes than relatively lower ones. 

\begin{figure}[t!]
    \begin{center}
       \includegraphics[width=0.85\linewidth]{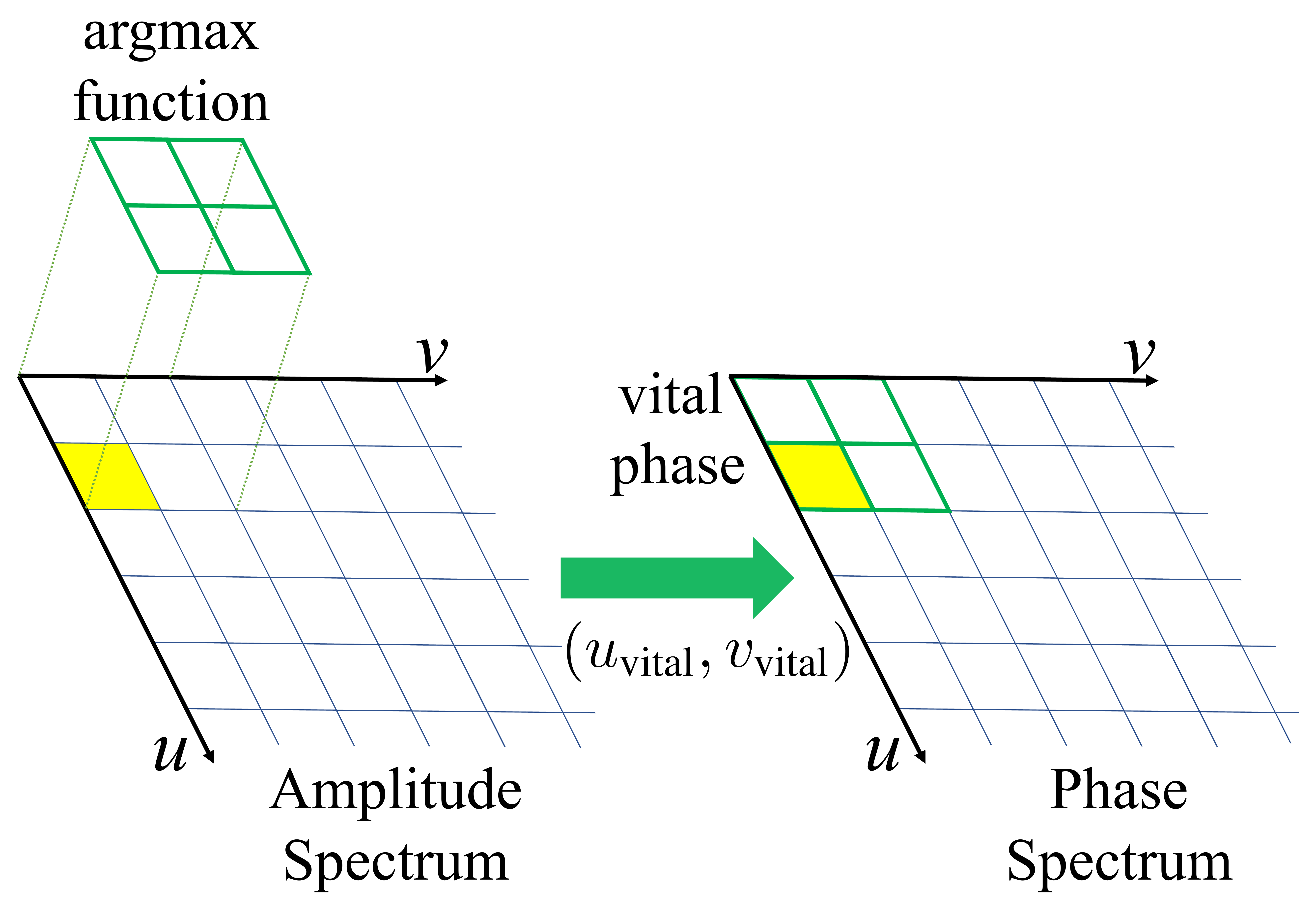}
    \end{center}
    \caption{Vital phase detection method. The vital phase coordinates are found by applying the argmax function to the filter size of the amplitude spectrum region. The filter moves on the amplitude spectrums without overlaps. This figure is shown in the case of a gray scale image.} 
    \label{figure4}
\end{figure}

The vital phase coordinates $(u_{\text{vital}},v_{\text{vital}},w_{\text{vital}})$ are determined by applying an $S \times S \times 1$ argmax filter to the amplitudes of each region as shown in Figure~\ref{figure4}. The filter encompasses all regions without any overlap. For sets $\mathcal{U}_{\text {vital }}$, $\mathcal{V}_{\text {vital }}$, and $\mathcal{W}_{\text {vital }}$ consisting of the elements $u_{\text{vital}}$, $v_{\text{vital}}$, and $w_{\text{vital}}$, respectively,
we get vital phase coordinate set $\mathcal{C}_{\text{vital}}$, where $\mathcal{C}_{\text{vital}}=\left\{(u, v, w) \mid u \in \mathcal{U}_{\text {vital }}, v \in \mathcal{V}_{\text {vital }}, w \in \mathcal{W}_{\text {vital }}\right\}$. 

We denote the vital phases ${P_{\text {vital }}(u, v, w)}$ and non-vital phases ${P_{\text {nonvital }}(u, v, w)}$ as follows:
\begin{equation} 
P(u,v,w)=
\begin{cases}
    P_{\text{vital}}(u,v,w) & \text{if } (u,v,w)\in \mathcal{C}_{\text{vital}} \\
    P_{\text{nonvital}}(u,v,w) & \text{otherwise}.
\end{cases}
\end{equation}
$P_{\text {vital }}(u,v,w)$ and  $P_{\text {nonvital }}(u,v,w)$ are classified based on robustness-related weights within a specified frequency range. The filter with a specific frequency range can prevent an increase in phase feature loss in that frequency range by not applying the filter to the entire phases at once.

\subsection{Vital Phase Augmentation}
To ensure that the model depends on the vital phases, we apply weak variations to the vital phases, whereas strong variations are applied to the non-vital phases. Excessive variations impede the model from learning domain-invariant features from the phase spectrum. There are two types of phase augmentation: vital phase augmentation using a Gaussian distribution (VIPAug-G) and using fractal phases (VIPAug-F)

\subsubsection{VIPAug-G.} VIPAug-G involves random sampling from a zero-mean Gaussian distribution and adding the obtained values to the phases. A Gaussian distribution exhibits high probability density around the mean and low probability density away from the mean. A Gaussian distribution effectively introduces finite variations to the phases to strengthen the model's dependency on the vital phases. The variations from Gaussian distributions $ \mathcal{N}$ with different variances are applied based on the corresponding weights:
\begin{equation}
P_{\text{aug}}^{\text {gauss }}(u, v, w)=P(u, v, w) + V(u,v,w),
\end{equation}
where a random variable $V(u,v,w)\sim  \mathcal{N}\left(0, \sigma_{\text {vital }}^2\right)$ if $(u,v,w)\in \mathcal{C}_{\text{vital}}$, $V(u,v,w) \sim  \mathcal{N}\left(0, \sigma_{\text {nonvital }}^2\right)$ otherwise and $\sigma_{\text {vital }}^2 \ll \sigma_{\text {nonvital }}^2$. 
With $-\pi \leq P_{\text {vital }} \leq \pi$ and $-\pi \leq P_{\text {nonvital }} \leq \pi$, the variation should be correspondingly small for the narrow range of vital and non-vital phases. In contrast to pixel-level perturbation caused by Gaussian noise, VIPAug-G introduces variations to the phases of the complex exponential functions that comprise an image by linear combination. 

\subsubsection{VIPAug-F.} VIPAug-F preserves the vital phases and entirely substitutes the non-vital phases with fractal phases. This method induces larger variations than VIPAug-G to enhance robustness against more significant fluctuations in phases. The replacement images should be from another domain and that domain should have different classes compared with the source domain. Replacing the original phases with those from other images from the same source domain prevents the model from depending on the phases, and diminishes the model's capability to learn domain-invariant features from the phases. Hence, we use the phases of a fractal image to enhance the structural complexity~\cite{hendrycks2022pixmix14} of the image. 
Fractal images are randomly chosen from a pool of 14,200 images. The non-vital phases are replaced with the fractal phases:
\begin{equation}
P_{\text{aug}}^{\text{frac}}(u,v,w)=
\begin{cases}
    P(u,v,w) & \text{if } (u,v,w)\in \mathcal{C}_{\text{vital}} \\
    P_{\text{fractal}}(u,v,w) & \text{otherwise},
\end{cases}
\end{equation}
where $P_{\text {fractal }}(u,v,w)$ denotes phases of the fractal image. 

VIPAug-F is designed to be robust against stronger phase fluctuations. However, completely replacing the non-vital phases may result in a substantial loss of image features. To retain original image features in the non-vital phases, we randomly apply VIPAug-F each iteration. Additionally, optional modifications may be necessary according to the training dataset. Due to more image features falling into the low-frequency region~\cite{li2023robust24}, the non-vital phase with the highest weight is retained at the low-frequency region. This is because the non-vital phases also have different relative robustness weights depending on the magnitude of the amplitudes.

\subsubsection{VIPAug.}
VIPAug combines amplitude augmentation and two types of phase augmentation. Denoting VIPAug-G as function $g(\cdot)$ and VIPAug-F as function $h(\cdot)$, we obtain
\begin{equation}
P_{\text {aug}} = g\circ t\circ h(P),
\end{equation}
where $\circ$ denotes function composition and $t(\cdot)$ is a phase change by pixel-wise augmentations from AutoAugment. The augmented amplitude $A_{\text {aug}}$ is obtained by APR-SP. We can reconstruct the augmented image $f_{\text{aug}}$ through iDFT with $P_{\text {aug}}$ and $A_{\text {aug}}$.

\begin{table}[!tbp]
\begin{center}
    {
        \renewcommand{\tabcolsep}{1.7mm}
        \begin{tabular}{c|c|cc}
        \toprule
        Dataset                    & Method   & \begin{tabular}[c]{@{}c@{}} \mc{Clean} \\ \mc{Acc (\%)}\end{tabular} & \begin{tabular}[c]{@{}c@{}} \mc{Corruption} \\ \mc{error rate (\%)} \end{tabular} \\ 
        \midrule
        \multirow{6}{*}{CIFAR-10}  & Baseline & 95.3                                                 & 25.3                                                             \\ 
                                   & APR-SP   & 95.6                                                 & 8.7                                                              \\ 
                                   & PRIME    & 93.9                                                 & 10.1                                                             \\ 
                                   & VIPAug-G & 95.6                                                 & \textbf{8.4}                                                     \\ 
                                   & VIPAug-F & \textbf{95.8}                                        & 8.6                                                              \\ 
                                   & VIPAug   & \textbf{95.8}                                        & \textbf{8.4}                                                     \\ 
                                   \midrule
        \multirow{6}{*}{CIFAR-100} & Baseline & \textbf{78.3}                                                 & 51.3                                   \\ 
                                   & APR-SP   & 76.7                                                 & 31.5                                                             \\ 
                                   & PRIME    & 76.8                                                 & 31.9                                                             \\ 
                                   & VIPAug-G & 76.9                                                 & \textbf{31.2}                                                    \\ 
                                   & VIPAug-F & 77.3                                        & 31.3                                                             \\ 
                                   & VIPAug   & 77.2                                                 & \textbf{31.2}                                                    \\ 
                                   \bottomrule 
        \end{tabular}
        }
\end{center}
\caption{Comparison with state-of-the-art methods on CIFAR-10 and CIFAR-100. The clean accuracy and corruption error rate were evaluated.}
\label{table1}
\end{table}

\section{Experiments}
\subsubsection{Datasets.}
We experimentally evaluated the performance of VIPAug on the most widely used CIFAR-10, CIFAR-100, ImageNet-100, and ImageNet datasets. CIFAR-10 and CIFAR-100 comprise 50,000 training images and 10,000 testing images, and each image is a 32$\times$32 color image with 10 classes and 100 classes, respectively. ImageNet consists of 1.2 million images and 1,000 classes. ImageNet-100 consists of 100 randomly selected classes of ImageNet. The training and test dataset contain 1,300 and 50 images per class, respectively. We used 14,200 fractal images from collections on DeviantArt to train the model. To measure the domain generalization performance, we used the corrupted datasets CIFAR-10-C, CIFAR-100-C, ImageNet-100-C, and ImageNet-C~\cite{hendrycks2019benchmarking27}, which contain 15 types of corruption, including noise, blur, weather, and digital corruption. Each type is demonstrated at five levels of severity.

\subsubsection{Metrics.}
We evaluated the domain generalization performance of the proposed method by measuring its accuracy on clean images and classification error rates on corrupted images. We also used the mean corruption error (mCE) on ImageNet-100-C and ImageNet-C, which is a normalized measure of the classification error rate by using AlexNet~\cite{krizhevsky2012imagenet1}. The corrupted test data has five severity levels $1 \leq s \leq 5$. The corruption error for each type of corruption was calculated as follows: $\mathrm{CE}_{\text {Corruption }}^{\text {Network }}=\sum_{s=1}^5 E_{s,\text{Corruption}}^{\text {Network }}/ \sum_{s=1}^5 E_{s,\text{Corruption}}^{\text {AlexNet }}$. We then calculate the mCE by averaging the $\mathrm{CE}_{\text {Corruption }}^{\text {Network }}$ for each type of corruption.

\subsection{CIFAR-10 and CIFAR-100 }
\subsubsection{Training Setup.}
We used a ResNet-18~\cite{he2016deep28} architecture as a baseline model. We trained all methods for 250 epochs. Detailed training setup can be seen in the supplementary material. We used the $2 \times 2 \times 1$ argmax filter, and set $\sigma_{\text {vital }}=0.001$ and $\sigma_{\text {nonvital }}=0.014$ on CIFAR-10 and $\sigma_{\text {vital }}=0.005$ and $\sigma_{\text {nonvital }}=0.012$ on CIFAR-100. VIPAug-G uses small values for the variance to introduce small variation to the phase. More details can be seen in the supplementary material. We also applied the modification to VIPAug-F on CIFAR-10 by setting the low-frequency region to 4/9 of the total phase. The non-vital phase with the highest weight is retained at the low frequency region.

\subsubsection{Results.}
\begin{table}[!tbp]
\begin{center}
    {
        \renewcommand{\tabcolsep}{1.5mm}
        \begin{tabular}{c|ccc}
        \toprule
        Method                                                       & \begin{tabular}[c]{@{}c@{}}Clean \\ Acc (\%) \end{tabular} & \begin{tabular}[c]{@{}c@{}}Corruption \\ error rate (\%) \end{tabular} & mCE (\%)           \\ 
        \midrule
        Baseline                                                     & 81.8                                                 & 52.8                                                             & 82.6          \\ 
        APR-SP                                                       & 82.2                                                 & 41.0                                                             & 65.4          \\ 
        PRIME                                                        & 80.2                                                 & 38.3                                                             & 60.9          \\ 
        VIPAug-G                                                     & 82.1                                                 & 39.5                                                             & 63.4          \\ 
         VIPAug-F                                                     & \textbf{82.4}                                        & 40.2                                                             & 64.5          \\ 
        VIPAug                                                       & 82.3                                                 & 39.4                                                             & 63.2          \\ 
        VIPAug with color & 80.5                                                 & 38.3                                                             & 61.0          \\ 
        VIPAug + PRIME & 79.8                                                 & \textbf{34.3}                                                    & \textbf{55.4} \\ \bottomrule
        \end{tabular}
    }
\end{center}
\caption{Comparison with state-of-the-art methods on ImageNet-100. Clean accuracy, corruption error rate, and mean corruption error (mCE) were evaluated.}
\label{table2}
\end{table}

\begin{table*}[!tbp]
\begin{center}
\begin{adjustbox}{max width=0.996\textwidth}
\renewcommand{\tabcolsep}{0.7mm}
{\Huge
\begin{tabular}{cc|ccc|cccc|cccc|cccc|c}
\toprule[1.8pt]
\multicolumn{2}{c|}{}                         & \multicolumn{3}{c|}{Noise} & \multicolumn{4}{c|}{Blur}       & \multicolumn{4}{c|}{Weather} & \multicolumn{4}{c|}{Digital}      &      \\ 
\midrule
\multicolumn{1}{c|}{Method}           & Clean & Gauss.  & Shot  & Impulse & Defocus & Glass & Motion & Zoom & Snow  & Frost & Fog & Bright & Contrast & Elastic & Pixel & JPEG & mCE  \\ 
\midrule
\multicolumn{1}{c|}{Baseline}         & 23.9  & 79      & 80    & 82       & 82      & 90    & 84     & 80   & 86    & 81    & 75  & 65     & 79       & 91      & 77    & 80   & 80.6 \\ 
\multicolumn{1}{c|}{Patch Uniform}    & 24.5  & 67      & 68    & 70       & 74      & 83    & 81     & 77   & 80    & 74    & 75  & 62     & 77       & 84      & 71    & 71   & 74.3 \\ 
\multicolumn{1}{c|}{AutoAug (AA)} & 22.8  & 69      & 68    & 72       & 77      & 83    & 80     & 81   & 79    & 75    & 64  & 56     & 70       & 88      & 57    & 71   & 72.7 \\ 
\multicolumn{1}{c|}{Random AA}        & 23.6  & 70      & 71    & 72       & 80      & 86    & 82     & 81   & 81    & 77    & 72  & 61     & 75       & 88      & 73    & 72   & 76.1 \\ 
\multicolumn{1}{c|}{MaxBlur pool}    & 23.0  & 73      & 74    & 76       & 74      & 86    & 78     & 77   & 77    & 72    & 63  & 56     & 68       & 86      & 71    & 71   & 73.4 \\ 
\multicolumn{1}{c|}{SIN}              & 27.2  & 69      & 70    & 70       & 77      & 84    & 76     & 82   & 74    & 75    & 69  & 65     & 69       & 80      & 64    & 77   & 73.3 \\ 
\multicolumn{1}{c|}{AugMix}           & \textbf{22.4}  & 65      & 66    & 67       & \textbf{70}      & \textbf{80}    & \textbf{66}     & \textbf{66}   & 75    & 72    & 67  & 58     & \textbf{58}       & \textbf{79}      & 69    & 69   & 68.4 \\ 
\multicolumn{1}{c|}{APR-SP}           & 24.4  & 60      & 64    & 63       & \textbf{70}      & 85    & 69     & 80   & 68    & 68    & 56  & 56     & 63       & 81      & 65    & \textbf{63}   & 67.4 \\ 
\multicolumn{1}{c|}{VIPAug (Ours)}    & 24.1  & \textbf{56}      & \textbf{59}    & \textbf{57}       & \textbf{70}      & 84    & 69     & 79   & \textbf{64}    & \textbf{64}    & \textbf{55}  & \textbf{55}     & 65       & 81      & \textbf{63}    & 67   & \textbf{65.8} \\
\bottomrule[1.8pt]
\end{tabular}%
}
\end{adjustbox}
\end{center}
\caption{Comparison with state-of-the-art methods on ImageNet dataset. For each corruption type, the average performance was evaluated. The mean corruption error (mCE) is a normalized average measure of the classification error rate on 15 corruptions.}
\label{table3}
\end{table*}

Table~\ref{table1} shows the performance comparison with the state-of-the-art models on CIFAR-10 and CIFAR-100. The baseline model achieved 95.3\% accuracy on the clean domain of CIFAR-10. However, the performance dropped significantly with an error rate of 25.3\% in the corrupted domain, which shows the importance of domain generalization. PRIME improved performance in the corrupted domain with primitive augmentations, but the method suffered from performance degradation in the clean domain. APR-SP fixes the phases and replaces the amplitudes with the amplitudes of other images. APR-SP improved accuracy in both the clean and corrupted domains. However, the method only considers the amplitude replacement, making it vulnerable to phase fluctuations in common corruptions. 

VIPAug combines phase variations with amplitude replacement to perform robustly against phase and amplitude fluctuations. VIPAug achieved an accuracy of $95.8 \%$ and a corruption error rate of $8.4\%$. Compared with APR-SP, these values were $0.2\%$p and $0.3\%$p better, respectively. VIPAug outperformed all the other methods on both clean and corrupted datasets. VIPAug-G and VIPAug achieved the lowest error rate for corrupted data, while VIPAug-F and VIPAug achieved the highest accuracy for clean data. 

Compared with APR-SP, VIPAug-G outperformed on the corrupted domain, indicating that VIPAug-G is more robust to corruption without sacrificing accuracy on uncorrupted data. VIPAug-F also exhibited improved performance for clean and corrupted data compared with APR-SP, despite the strong variation. These results show that the model still learns domain-invariant features from the phase, even with the high variation of VIPAug-F. VIPAug achieved state-of-the-art performance on CIFAR-10 and CIFAR-100.

\subsection{ImageNet-100 and ImageNet}
\subsubsection{Training Setup.}
We used a ResNet-18 architecture as the baseline model on ImageNet-100 and a ResNet-50~\cite{he2016deep28} model on ImageNet. The models were trained for 100 epochs. Detailed training setup can be seen in the supplementary material. We evaluated all methods on ImageNet-100 using the same training settings. For ImageNet, we used pretrained weights for alternative methods if available. Otherwise, we used the performance results reported in AugMix. We used the $2\times2\times1$ argmax filter, and set $\sigma_{\text {vital }}=0.001$ and $\sigma_{\text {nonvital }}=0.005$. We applied the modification to VIPAug-F by setting the low-frequency region to 1/4 of the total phase. 

\subsubsection{Results.}
In Table~\ref{table2}, we compared VIPAug with APR-SP and PRIME on ImageNet-100. The variations of VIPAug exhibited greater accuracy on the clean domain compared with the other methods. VIPAug also showed greater clean accuracy by $0.1\%$p and decreased mCE by $2.2\%$p compared with APR-SP. PRIME reduced the clean accuracy compared with the baseline, but significantly improved the generalization ability on the corrupted domain. We conjecture that the diverse color transformation of PRIME contributed to the generalization capability. We added the color transformation of PRIME to VIPAug and compared the performance. VIPAug with color decreased the mCE by $4.4 \%$p compared with APR-SP and achieved nearly state-of-the-art performance compared with PRIME. When VIPAug and PRIME were applied together, they showed an overwhelming performance of $55.4 \%$ mCE. These results confirm that VIPAug and PRIME can be considered somewhat orthogonal approaches.

In Table~\ref{table3}, domain generalization methods were evaluated on ImageNet for each corruption type~\cite{lopes2019patch29,autoaugment13,zhang2019maxblur31,rusak2020sin32,augmix17,chen2021amplitude15}. VIPAug did not excel in all corruption types, but the method achieved the best average performance on the corrupted domains while maintaining performance on the clean domain. This demonstrates the effectiveness of VIPAug on a large-scale dataset.

\subsection{Ablation Studies}
\subsubsection{Robustness Weight.}

\begin{table}[!tbp]
\begin{center}
    { 
        \renewcommand{\tabcolsep}{1.2mm}
        \begin{tabular}{c|ccc}
        \toprule
        Method & \begin{tabular}[c]{@{}c@{}}Clean \\ Acc (\%)\end{tabular} & \begin{tabular}[c]{@{}c@{}}Corruption \\ error rate (\%) \end{tabular} & mCE (\%) \\ 
        \midrule
        APR-SP & 82.2 & 41.0  & 65.4 \\ 
        VIPAug & \textbf{82.3}  & \textbf{39.4}  & \textbf{63.2} \\ 
        Reverse VIPAug & 81.8  & 42.8  & 68.2 \\
        Uniform VIPAug (a) & 82.2  & 40.5  & 64.9 \\ 
        Uniform VIPAug (b) & 81.7 & 41.2  & 66.1 \\ 
        \bottomrule
        \end{tabular}
    }
\end{center}
\caption{The ablation analysis of robustness weight to verify two hypotheses. VIPAug outperformed Reverse VIPAug, implying that vital phase contains more robust features. Furthermore, VIPAug outperformed Uniform VIPAug (a) and (b), implying that strengths of variation should be proportional to the robustness of each phase.}
\label{table4}
\end{table}

We evaluated our first hypothesis that vital phase contains more domain-invariant features than non-vital phase by comparing the performance of Reverse VIPAug and VIPAug. Reverse VIPAug treated vital phase as non-vital phase and one of non-vital phases as vital phase. If the robustness weights of vital phase and non-vital phase are the same, the performance should be similar. However, as shown in Table~\ref{table4}, Reverse VIPAug performed worse than VIPAug. The clean accuracy of Reverse VIPAug was 0.5\%p lower than that of VIPAug, and the mCE was 5.0\%p higher. This indicates that Reverse VIPAug is not as robust to corruption as VIPAug. In particular, when compared with APR-SP without phase variation, the clean accuracy of Reverse VIPAug was 0.4\%p lower and the mCE was 2.8\%p higher. This suggests that adding variation to the phase does not always improve performance.

We then evaluated the second hypothesis that the strengths of variation should be proportional to the robustness weight of each phase by comparing the performance of Uniform VIPAug (a) and Uniform VIPAug (b). Uniform VIPAug (a) set $\sigma_{\text {vital }}=0.001$ and $\sigma_{\text {nonvital }}=0.001$, and randomly replaced vital phase and non-vital phase with fractals to give the same strengths of variation. Uniform VIPAug (b) set $\sigma_{\text {vital }}=0.005$ and $\sigma_{\text {nonvital }}=0.005$, with other conditions same as Uniform VIPAug (a). Uniform VIPAug (a) and (b) exhibited lower performance on both clean and corrupted images compared with VIPAug. Our results suggest that the model becomes more robust to corruption if the strengths of variations are varied according to the robustness weights.

\begin{figure}[!tbp]
\centering
\includegraphics[width=0.9\columnwidth]{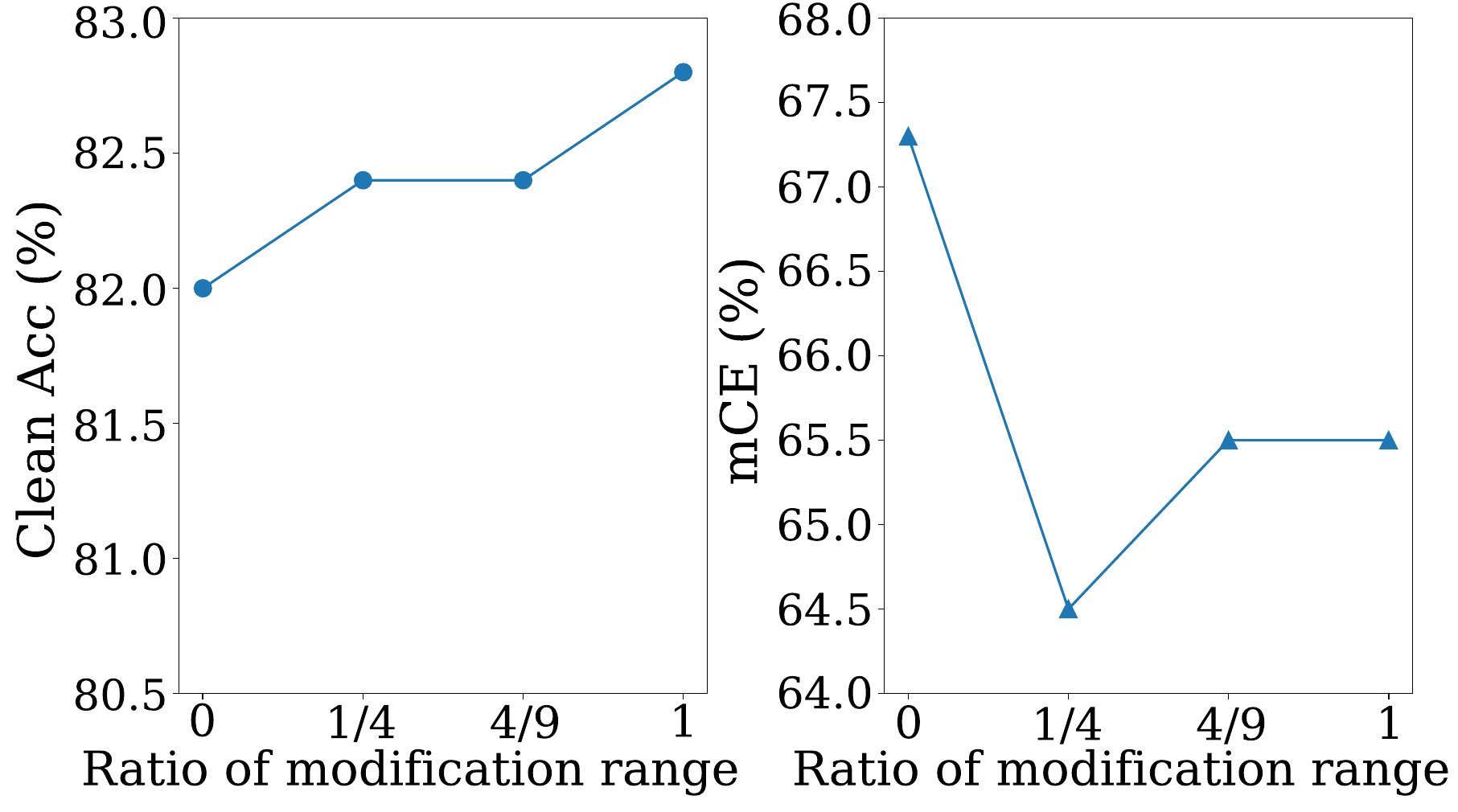}
\caption{The ablation analysis of modifications to the phase range of VIPAug-F on ImageNet-100 and ImageNet-100-C. The x-axis stands for the ratio of the modified phase range to the total phase spectrum.}
\label{fig5}
\end{figure}

\subsubsection{Modification on VIPAug-F.}
VIPAug-F introduces strong variations to partially replace phases with fractal images phases. To investigate the effects of different modification ranges on the performance of VIPAug-F, we conducted an ablation study on ImageNet-100.

Because the robustness weights are relatively different between non-vital phases, we modified VIPAug-F not to additionally replace the case with the second largest magnitude of amplitude at the low-frequency spectrum. This is because the low-frequency region contains more image features~\cite{li2023robust24}.

We conducted the experiment in four cases: no modification, modification of 1/4 of the entire phases, modification of 4/9, and modification of the entire phases, as shown in Figure~\ref{fig5}. We found that VIPAug-F with no modification had the lowest clean accuracy and the highest mCE. VIPAug-F with more modification had higher clean accuracy but lower mCE. These results suggest that finite variation should be given to the phase in order to improve the performance of VIPAug-F empirically. We need to find the appropriate hyperparameters that balance the clean accuracy and mCE.

\begin{table}[!tbp]
\begin{center}
    {
        \renewcommand{\tabcolsep}{1mm}
        \begin{tabular}{ccc}
        \toprule
        Dataset           & Clean Acc (\%) & Corruption error rate (\%)\\ 
        \midrule
        Fractal           & \textbf{77.3}      & \textbf{31.4}                  \\ 
        ImageNet (IN)          & 77.0      & 31.6                  \\ 
        Stylized-IN & 77.1      & 31.7                  \\ 
        GTA5              & 76.9      & 31.5                  \\
        \bottomrule
        \end{tabular}
     }
\end{center}
\caption{Ablation analysis of VIPAug-F for other datasets. Only fractal dataset has no class.}
\label{table5}
\end{table}

\subsubsection{Other Datasets for VIPAug-F.}

In Table~\ref{table5}, we compared the performance of VIPAug-F when using different datasets instead of fractals. We used ImageNet, Stylized-ImageNet, and GTA5~\cite{modas2022gta23} datasets. The number of images was 14,200, the same as fractal images. The 14,200 images were randomly selected from each dataset. Other datasets show a slight performance decrease for clean and corrupted datasets compared with fractal dataset. This is because ImageNet, Stylized-ImageNet, and GTA5 all have similar classes to the CIFAR-100. If two images of different classes are mixed together, the model cannot depend on the phase well. On the other hand, the fractal images have no class. Fractals also introduce structural complexity to images~\cite{hendrycks2022pixmix14}. CIFAR-100 has a wide variety of classes, making it difficult to find a dataset consisting of completely different images. Therefore, the dataset without a class is more suitable.

\subsubsection{Comparison with 2D DFT.}
\begin{figure}[!tbp]
\centering
\includegraphics[width=0.9\columnwidth]{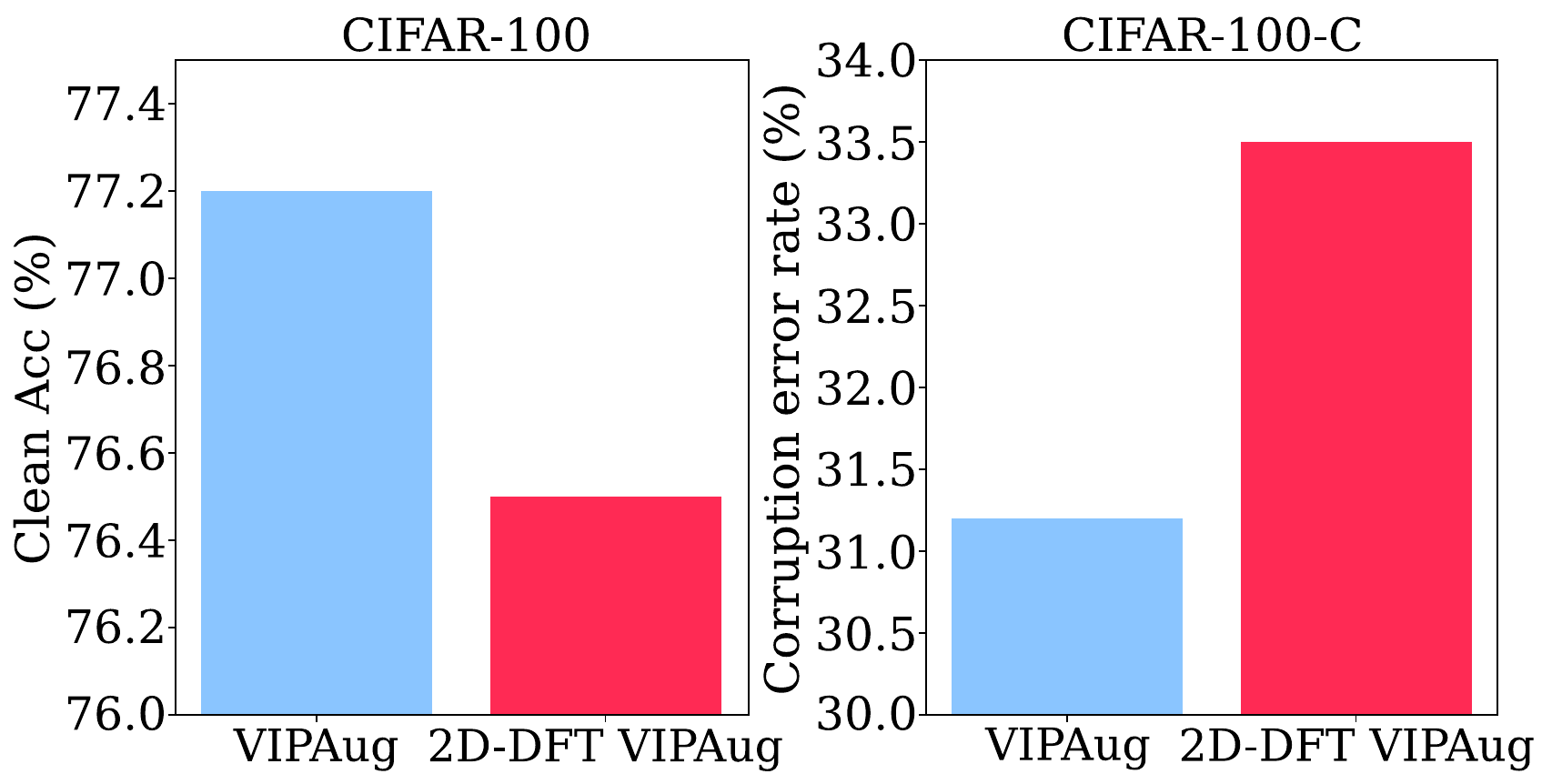}
\caption{Comparison of applying 2D DFT and 3D DFT to VIPAug on CIFAR-100.}
\label{figure6}
\end{figure}

We compared the performance of VIPAug and 2D-DFT VIPAug in Figure~\ref{figure6}. 2D-DFT VIPAug showed 0.7\%p lower clean accuracy and 2.3\%p higher corruption error rate than VIPAug. Extending along the channel axis in 3D DFT allows vital phases to be identified for all channels when the filter is applied. Considering the large performance difference, amplitude and phase features between channels have a significant impact on the model's performance.

\section{Conclusion}
We first argue that the robustness weight differs for each phase of the image. We propose a novel method to classify vital and non-vital phases according to their weights. We make the model more robust to corruption by giving different strengths of variation to the phases according to their weights. Our extensive experimental results showed that our approach achieved SOTA performance on CIFAR-10 and CIFAR-100, and achieved performance close to the SOTA methods on ImageNet-100 and ImageNet. In this study, we presented a new perspective on the phase in domain generalization research. We suggested a promising direction for subsequent research on how to deal with the image phase.

\section{Acknowledgments}
This work was supported in part by Institute of Information$\And$communications Technology Planning$\And$Evaluation (IITP) grant funded by Korea government (MSIT) (No.2020-0-00440, Development of Artificial Intelligence Technology that Continuously Improves Itself as the Situation Changes in the Real World). This research was supported in part by the KAIST Convergence Research Institute Operation Program. The students are supported by the BK21 FOUR from the Ministry of Education (Republic of Korea).

\bibliography{aaai24}

\begin{thebibliography}{29}
\providecommand{\natexlab}[1]{#1}

\bibitem[{Chen et~al.(2021)Chen, Peng, Ma, Li, Du, and Tian}]{chen2021amplitude15}
Chen, G.; Peng, P.; Ma, L.; Li, J.; Du, L.; and Tian, Y. 2021.
\newblock {{Amplitude-phase recombination: Rethinking robustness of convolutional neural networks in frequency domain}}.
\newblock In \emph{Proceedings of the IEEE/CVF International Conference on Computer Vision}, 458--467.

\bibitem[{Cubuk et~al.(2019)Cubuk, Zoph, Mane, Vasudevan, and Le}]{autoaugment13}
Cubuk, E.~D.; Zoph, B.; Mane, D.; Vasudevan, V.; and Le, Q.~V. 2019.
\newblock {AutoAugment: Learning augmentation policies from data}.
\newblock In \emph{Proceedings of the IEEE/CVF Conference on Computer Vision and Pattern Recognition}, 113--123.

\bibitem[{Deng et~al.(2009)Deng, Dong, Socher, Li, Li, and Fei-Fei}]{deng2009imagenet26}
Deng, J.; Dong, W.; Socher, R.; Li, L.-J.; Li, K.; and Fei-Fei, L. 2009.
\newblock {{ImageNet: A large-scale hierarchical image database}}.
\newblock In \emph{Proceedings of the IEEE/CVF Conference on Computer Vision and Pattern Recognition}, 248--255.

\bibitem[{DeVries and Taylor(2017)}]{cutout20}
DeVries, T.; and Taylor, G.~W. 2017.
\newblock Improved regularization of convolutional neural networks with cutout.
\newblock arXiv:1708.04552.

\bibitem[{Ding and Fu(2017)}]{ding2017deep5}
Ding, Z.; and Fu, Y. 2017.
\newblock {{Deep domain generalization with structured low-rank constraint}}.
\newblock \emph{IEEE Transactions on Image Processing}, 27(1): 304--313.

\bibitem[{Dosovitskiy et~al.(2021)Dosovitskiy, Beyer, Kolesnikov, Weissenborn, Zhai, Unterthiner, Dehghani, Minderer, Heigold, Gelly, Uszkoreit, and Houlsby}]{dosovitskiy2021an}
Dosovitskiy, A.; Beyer, L.; Kolesnikov, A.; Weissenborn, D.; Zhai, X.; Unterthiner, T.; Dehghani, M.; Minderer, M.; Heigold, G.; Gelly, S.; Uszkoreit, J.; and Houlsby, N. 2021.
\newblock {An image is worth 16x16 words: Transformers for image recognition at scale}.
\newblock In \emph{International Conference on Learning Representations}.

\bibitem[{He et~al.(2016)He, Zhang, Ren, and Sun}]{he2016deep28}
He, K.; Zhang, X.; Ren, S.; and Sun, J. 2016.
\newblock {Deep residual learning for image recognition}.
\newblock In \emph{Proceedings of the IEEE/CVF Conference on Computer Vision and Pattern Recognition}, 770--778.

\bibitem[{Hendrycks and Dietterich(2019)}]{hendrycks2019benchmarking27}
Hendrycks, D.; and Dietterich, T. 2019.
\newblock {Benchmarking neural network robustness to common corruptions and perturbations}.
\newblock In \emph{International Conference on Learning Representations}.

\bibitem[{Hendrycks et~al.(2019)Hendrycks, Mu, Cubuk, Zoph, Gilmer, and Lakshminarayanan}]{augmix17}
Hendrycks, D.; Mu, N.; Cubuk, E.~D.; Zoph, B.; Gilmer, J.; and Lakshminarayanan, B. 2019.
\newblock {{AugMix: A simple data processing method to improve robustness and Uncertainty}}.
\newblock In \emph{International Conference on Learning Representations}.

\bibitem[{Hendrycks et~al.(2022)Hendrycks, Zou, Mazeika, Tang, Li, Song, and Steinhardt}]{hendrycks2022pixmix14}
Hendrycks, D.; Zou, A.; Mazeika, M.; Tang, L.; Li, B.; Song, D.; and Steinhardt, J. 2022.
\newblock {{PixMix: Dreamlike pictures comprehensively improve safety measures}}.
\newblock In \emph{Proceedings of the IEEE/CVF International Conference on Computer Vision}, 16783--16792.

\bibitem[{Krizhevsky and Hinton(2009)}]{krizhevsky2009cifar25}
Krizhevsky, A.; and Hinton, G. 2009.
\newblock {Learning multiple layers of features from tiny images}.
\newblock Technical Report TR-2009, Dept.\ of Computer Science, Toronto Univ.

\bibitem[{Krizhevsky, Sutskever, and Hinton(2012)}]{krizhevsky2012imagenet1}
Krizhevsky, A.; Sutskever, I.; and Hinton, G.~E. 2012.
\newblock {{ImageNet classification with deep convolutional neural networks}}.
\newblock \emph{Advances in Neural Information Processing Systems}, 25.

\bibitem[{Lee and Myung(2022)}]{lee2022adversarial}
Lee, W.; and Myung, H. 2022.
\newblock {Adversarial attack for asynchronous event-based data}.
\newblock In \emph{Proceedings of the AAAI Conference on Artificial Intelligence}, volume~36, 1237--1244.

\bibitem[{Li et~al.(2023)Li, Ortega~Caro, Rusak, Brendel, Bethge, Anselmi, Patel, Tolias, and Pitkow}]{li2023robust24}
Li, Z.; Ortega~Caro, J.; Rusak, E.; Brendel, W.; Bethge, M.; Anselmi, F.; Patel, A.~B.; Tolias, A.~S.; and Pitkow, X. 2023.
\newblock {{Robust deep learning object recognition models rely on low frequency information in natural images}}.
\newblock \emph{PLOS Computational Biology}, 19(3): e1010932.

\bibitem[{Liu et~al.(2020)Liu, Dou, Yu, and Heng}]{liu2020ms6}
Liu, Q.; Dou, Q.; Yu, L.; and Heng, P.~A. 2020.
\newblock {{MS-Net: multi-site network for improving prostate segmentation with heterogeneous MRI data}}.
\newblock \emph{IEEE Transactions on Medical Imaging}, 39(9): 2713--2724.

\bibitem[{Lopes et~al.(2019)Lopes, Yin, Poole, Gilmer, and Cubuk}]{lopes2019patch29}
Lopes, R.~G.; Yin, D.; Poole, B.; Gilmer, J.; and Cubuk, E.~D. 2019.
\newblock Improving robustness without sacrificing accuracy with patch gaussian augmentation.
\newblock arXiv:1906.02611.

\bibitem[{Modas et~al.(2022)Modas, Rade, Ortiz-Jim{\'e}nez, Moosavi-Dezfooli, and Frossard}]{modas2022prime18}
Modas, A.; Rade, R.; Ortiz-Jim{\'e}nez, G.; Moosavi-Dezfooli, S.-M.; and Frossard, P. 2022.
\newblock {PRIME: A few primitives can boost robustness to common corruptions}.
\newblock In \emph{Proceedings of the European Conference on Computer Vision}, 623--640. {Springer}.

\bibitem[{Motiian et~al.(2017)Motiian, Piccirilli, Adjeroh, and Doretto}]{motiian2017unified2}
Motiian, S.; Piccirilli, M.; Adjeroh, D.~A.; and Doretto, G. 2017.
\newblock {{Unified deep supervised domain adaptation and generalization}}.
\newblock In \emph{Proceedings of the IEEE/CVF International Conference on Computer Vision}, 5715--5725.

\bibitem[{Richter et~al.(2016)Richter, Vineet, Roth, and Koltun}]{modas2022gta23}
Richter, S.~R.; Vineet, V.; Roth, S.; and Koltun, V. 2016.
\newblock {Playing for data: Ground truth from computer games}.
\newblock In \emph{Proceedings of the European Conference on Computer Vision}, 102--118. {Springer}.

\bibitem[{Rusak et~al.(2020)Rusak, Schott, Zimmermann, Bitterwolf, Bringmann, Bethge, and Brendel}]{rusak2020sin32}
Rusak, E.; Schott, L.; Zimmermann, R.~S.; Bitterwolf, J.; Bringmann, O.; Bethge, M.; and Brendel, W. 2020.
\newblock {A simple way to make neural networks robust against diverse image corruptions}.
\newblock In \emph{Proceedings of the European Conference on Computer Vision}, 53--69. {Springer}.

\bibitem[{Tan and Le(2019)}]{tan2019efficientnet}
Tan, M.; and Le, Q. 2019.
\newblock {EfficientNet: Rethinking model scaling for convolutional neural networks}.
\newblock In \emph{International Conference on Machine Learning}, 6105--6114. PMLR.

\bibitem[{Xu et~al.(2021)Xu, Zhang, Zhang, Wang, and Tian}]{xu2021fourier16}
Xu, Q.; Zhang, R.; Zhang, Y.; Wang, Y.; and Tian, Q. 2021.
\newblock {A fourier-based framework for domain generalization}.
\newblock In \emph{Proceedings of the IEEE/CVF Conference on Computer Vision and Pattern Recognition}, 14383--14392.

\bibitem[{Yoon, Hamarneh, and Garbi(2019)}]{yoon2019generalizable3}
Yoon, C.; Hamarneh, G.; and Garbi, R. 2019.
\newblock {{Generalizable feature learning in the presence of data bias and domain class imbalance with application to skin lesion classification}}.
\newblock In \emph{International Conference on Medical Image Computing and Computer Assisted Intervention}, 365--373. {Springer}.

\bibitem[{Yucel, Cinbis, and Duygulu(2023)}]{hybridaug2023}
Yucel, M.~K.; Cinbis, R.~G.; and Duygulu, P. 2023.
\newblock {{HybridAugment++: Unified frequency spectra perturbations for model robustness}}.
\newblock In \emph{Proceedings of the IEEE/CVF International Conference on Computer Vision}, 5718--5728.

\bibitem[{Zhang et~al.(2018)Zhang, Cisse, Dauphin, and Lopez-Paz}]{mixup19}
Zhang, H.; Cisse, M.; Dauphin, Y.~N.; and Lopez-Paz, D. 2018.
\newblock {{Mixup: Beyond empirical risk minimization}}.
\newblock In \emph{International Conference on Learning Representations}.

\bibitem[{Zhang(2019)}]{zhang2019maxblur31}
Zhang, R. 2019.
\newblock {Making convolutional networks shift-invariant again}.
\newblock In \emph{International Conference on Machine Learning}, 7324--7334. {PMLR}.

\bibitem[{Zhao et~al.(2021)Zhao, Zhong, Yang, Luo, Lin, Li, and Sebe}]{zhao2021learning10}
Zhao, Y.; Zhong, Z.; Yang, F.; Luo, Z.; Lin, Y.; Li, S.; and Sebe, N. 2021.
\newblock {{Learning to generalize unseen domains via memory-based multi-source meta-learning for person re-identification}}.
\newblock In \emph{Proceedings of the IEEE/CVF International Conference on Computer Vision and Pattern Recognition}, 6277--6286.

\bibitem[{Zhong et~al.(2020)Zhong, Zheng, Kang, Li, and Yang}]{zhong2020random11}
Zhong, Z.; Zheng, L.; Kang, G.; Li, S.; and Yang, Y. 2020.
\newblock {Random erasing data augmentation}.
\newblock In \emph{Proceedings of the AAAI Conference on Artificial Intelligence}, volume~34, 13001--13008.

\bibitem[{Zhou et~al.(2021)Zhou, Yang, Qiao, and Xiang}]{mixstyle12}
Zhou, K.; Yang, Y.; Qiao, Y.; and Xiang, T. 2021.
\newblock {{Domain generalization with mixstyle}}.
\newblock In \emph{International Conference on Learning Representations}.

\end{thebibliography}

\end{document}